\documentclass[10pt,twocolumn,letterpaper]{article}

\usepackage[pagenumbers]{cvpr} % To force page numbers, e.g. for an arXiv version
\usepackage{pifont}
\usepackage{graphicx}
\usepackage{overpic}

\definecolor{cvprblue}{rgb}{0.21,0.49,0.74}
\usepackage[pagebackref,breaklinks,colorlinks,citecolor=cvprblue]{hyperref}
\usepackage{graphicx}
\usepackage{booktabs}
\usepackage[misc]{ifsym}
\usepackage{lipsum}  
\usepackage[accsupp]{axessibility}  % Improves PDF readability for those with disabilities.

\usepackage{listings}
\usepackage[ruled,linesnumbered]{algorithm2e}
\SetKwComment{Comment}{/* }{ */}
\usepackage{amsmath}  % add by shengxuhan
\usepackage{multicol}
\usepackage{multirow}
\usepackage{makecell}
\usepackage{adjustbox}
\usepackage{bbding}
\usepackage{appendix}
\usepackage{orcidlink}
\usepackage{amsfonts}       % blackboard math symbols
\usepackage{nicefrac}       % compact symbols for 1/2, etc.
\usepackage{microtype}      % microtypography
\usepackage{xcolor}         % colors
\usepackage {pifont}
\usepackage{subcaption}
\usepackage{colortbl}
\definecolor{lightgray}{gray}{0.95}
\definecolor{color3}{gray}{0.95}
\definecolor{rouse}{rgb}{0.981,0.961,0.941}
\usepackage{float}
\usepackage{wrapfig}
\usepackage{courier}
\usepackage[frozencache,cachedir=minted-cache]{minted}
\newcommand{\boldblue}[1]{\textcolor{blue}{\underline{#1}}}
\newcommand{\boldred}[1]{\textcolor{Red}{\textbf{#1}}}
\newcolumntype{T}{@{\hspace{0em}}c@{\hspace{0em}}}
\def\paperID{a} % *** Enter the Paper ID here
\def\confName{CVPR}
\def\confYear{2025}

\title{GaussianSeal: Rooting Adaptive Watermarks for 3D Gaussian Generation Model}

\author{
    Runyi Li\quad Xuanyu Zhang\quad Chuhan Tong\quad Zhipei Xu\quad Jian Zhang$^{\dagger}$\\
    School of Electronic and Computer Engineering, Peking University, Shenzhen, China
}
\begin{document}
\maketitle
\renewcommand*{\thefootnote}{$\dagger$}
\footnotetext[1]{Corresponding author.}
\begin{abstract}
With the advancement of AIGC technologies, the modalities generated by models have expanded from images and videos to 3D objects, leading to an increasing number of works focused on 3D Gaussian Splatting (3DGS) generative models. Existing research on copyright protection for generative models has primarily concentrated on watermarking in image and text modalities, with little exploration into the copyright protection of 3D object generative models. In this paper, we propose the first bit watermarking framework for 3DGS generative models, named GaussianSeal, to enable the decoding of bits as copyright identifiers from the rendered outputs of generated 3DGS. By incorporating adaptive bit modulation modules into the generative model and embedding them into the network blocks in an adaptive way, we achieve high-precision bit decoding with minimal training overhead while maintaining the fidelity of the model's outputs. Experiments demonstrate that our method outperforms post-processing watermarking approaches for 3DGS objects, achieving superior performance of watermark decoding accuracy and preserving the quality of the generated results.
\end{abstract}
\section{Introduction}
\label{sec:intro}
The advancement of AI Generated Content (AIGC) has marked a significant shift in how we perceive and interact with digital media. Over the years, generative models have evolved dramatically, enhancing their capabilities in producing high-quality images and videos~\cite{rombach2021highresolution,peebles2023scalable,ho2022imagen,singer2022make}. Recently, AIGC technologies have begun to extend beyond 2D images, venturing into the realm of 3D model generation. This transition not only broadens the applications of AIGC in fields such as gaming, virtual reality, and architecture but also poses new challenges and opportunities for researchers and practitioners in computer vision. As we delve into this emerging domain, recognizing the importance of safety considerations in AIGC models becomes essential for guiding future advancements in both generative modeling and the broader realm of computer vision~\cite{min2024uncovering,ci2024wmadapter,ma2024safe,peng2023intellectual,zhang2023editguard} and AI generation~\cite{zhang2024v2a,xu2024fakeshield}.

3D Gaussian Splatting (3DGS)~\cite{kerbl20233d} has emerged as a new generation of representation methods for 3D objects and scenes, providing efficient and high-quality 3D models~\cite{li2023generative,fei20243d}. Certain approaches utilize images as guiding inputs to generate high-quality 3D Gaussian models in a controllable manner~\cite{yi2023gaussiandreamer,liangluciddreamer2023,tang2025lgm,xu2024grm}. This versatility makes 3DGS particularly valuable in fields such as game development and interactive applications. As this technology evolves, it holds the potential to enhance the visual fidelity and interactivity of digital environments significantly.

% 动机is three-fold:(1) 现有模型水印没有做3d生成模型的，(2) 3D物体水印速度慢，in-generation相对post-generation, (3) 针对具体的模型水印方法，微调模型的approach会影响模型的生成质量，大模型难以在保留原有生成质量的同时完成高精度的水印嵌入。同时完成高精度的水印解码和高质量的3D生成是具有挑战性的。
\begin{figure}
    \centering
    \includegraphics[width=1\linewidth]{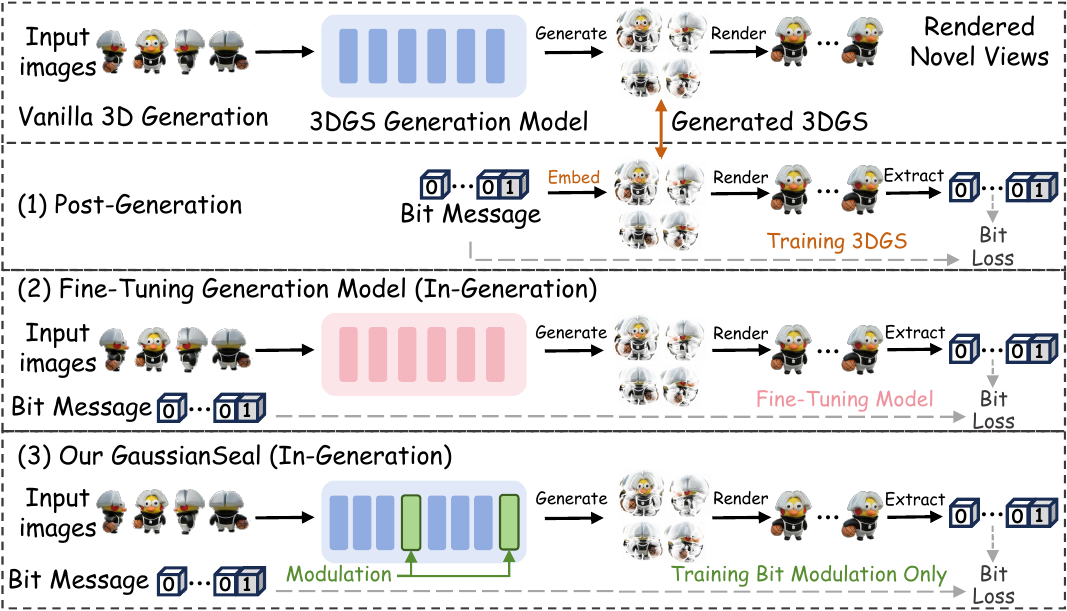}
    \caption{Current approaches for 3D Gaussian watermarking, including (1) post-generation methods, (2) fine-tuning generation model, and (3) our bit modulation into generation model. Our framework is lightweight, making trade-off between accurate bit extraction and keeping generation quality.}
    \label{fig:framework_compare}
    \vspace{-6mm}
\end{figure}
Current model watermarking methods~\cite{min2024watermark,rezaei2024lawa,feng2024aqualora,fernandez2023stable}, especially those designed for generative models, predominantly focus on text-to-image models, such as Stable Diffusion (SD)~\cite{rombach2021highresolution}. By fine-tuning or adding learnable adapter modules into the generation model, the model itself is able to be watermarked. While these techniques have made strides in embedding watermarks within image outputs, their application to 3DGS remains limited. 
There are also explorations addressing the protection of individual 3DGS objects~\cite{zhang2024gs,jang20243d,huang2024gaussianmarker,qi2024geometry}. With a fixed watermark decoder and the specially designed 3DGS structure, the 3DGS is trained with a watermark extracted via the decoder, but \textit{one at a time}. Illustrations of these methods and our proposed framework are shown in Fig.~\ref{fig:framework_compare}.

Our motivation is two-folds: \ding{172}~\textit{Post-generation} watermarking methods (Fig.~\ref{fig:framework_compare}(1)) only watermark one 3D object at a time, which is inefficient and \textbf{requires a significant amount of time} for each object; \ding{173}~For \textit{in-generation} watermarking methods, the approach of fine-tuning generative model (Fig.~\ref{fig:framework_compare}(2)) would \textbf{impact the generation quality}, which struggles to achieve precise watermark extraction while maintaining generation quality, and they \textbf{require large-scale datasets and substantial GPU memory}. In short, achieving both high-precision watermark decoding and high-quality 3D generation presents is challenging.

\begin{figure}
    \centering
    \includegraphics[width=1.0\linewidth]{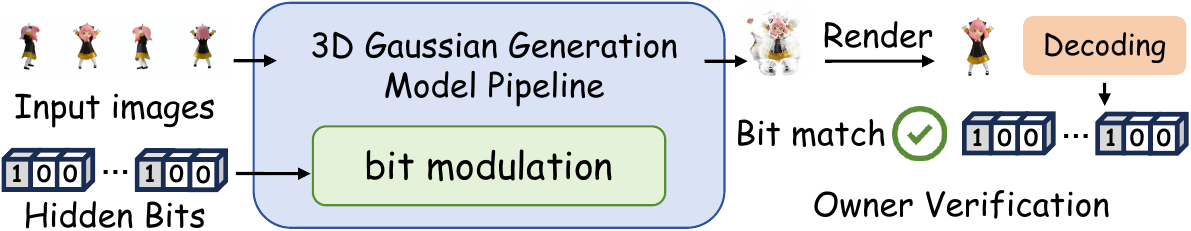}
    \caption{Application and watermarking process of our proposed framework GaussianSeal.}
    \label{fig:teaser}
    \vspace{-6mm}
\end{figure}

Motivated by the insights and limitations mentioned above, we propose the first bit watermarking framework for 3DGS generative models, taking the current state-of-the-art 3D generation model LGM~\cite{tang2025lgm} as an example. Specifically, we introduce an adaptive bit modulation mechanism that embeds secret messages directly into the model network (Fig.~\ref{fig:framework_compare}(3)). To ensure that the generative quality of the model remains unaffected, we only fine-tune the modulation module without altering the original model parameters. The added watermark can be extracted from the 3DGS renderings of LGM output, providing a robust method for owner verification and copyright protection in 3D content creation. The application scenarios and the overall watermarking process of our proposed method are shown in Fig.~\ref{fig:teaser}. Our contributions are listed as follows:

\noindent \ding{113}~We introduce the first bit watermarking framework for 3D generative models, enabling precise verification of bit information and effectively safeguarding the copyrights of 3DGS generating models.

\noindent \ding{113}~We present an adaptive and effective bit watermark modulation mechanism that achieves a balance between message decoding accuracy and the perceptual quality of the generative model, which exhibits strong robustness and security.

\noindent \ding{113}~Extensive experiments demonstrate that our watermarking framework delivers precise and robust performance in copyright protection for 3DGS generative models with time efficiency, achieving state-of-the-art results compared to existing post-generation methods and possible intuitive in-generation methods.

\section{Related Work}
\label{sec:relatedwork}

% \subsection{Model Watermarking}
% To verify the ownership of models and embed owner information, several watermarking techniques have been proposed, which are mainly divided into two types: one is embedding watermarks in the model's \textbf{weights}, and the other is embedding watermarks in the model's \textbf{output}. If the protectors of the model have an understanding of the model's structure, they can embed watermarks into the model's weights, and the owner can be identified by decoding these watermarks, and such methods are known as \textit{white-box watermarking}. Conversely, if the owner is unfamiliar with the model's structure, they can choose to embed watermarks in the model's output, such as by creating specific trigger sets and other techniques to claim ownership and such methods are known as \textit{black-box watermarking}.

% In certain special cases where the model structure is not explicitly defined, this type of situation is called \textit{box-free watermarking}. Zhang et al. proposed a method for image-to-image generation to embed special watermarks. The specific operation first adds watermarks to the images and then uses the model to process these images to produce outputs from which the watermarks can be extracted. For images that have not been processed by the model, a benign signal will be extracted. The method proposed in this study draws on this model-free (box-free) watermark design.

\subsection{Watermarking Generation Models}
With the continuous emergence of AIGC models, the traceability and copyright protection of AI-generated models have gradually attracted research interest. Most current watermarking methods for AI-generated models focus on Text-to-Image (T2I) models~\cite{zhang2023adding,mou2024t2i}, especially watermarking for Stable Diffusion~\cite{zhao2023recipe,liu2023watermarking,xiong2023flexible,lei2024diffusetrace,yuan2024watermarking,wang2024spy,kim2024wouaf,li2024protect}. Stable Signature~\cite{fernandez2023stable} proposed the first watermarking method for Stable Diffusion, which involves pre-training a watermark extractor applicable to both VAE~\cite{kingma2013auto} encoders and decoders, then fixing this extractor and fine-tuning the Stable Diffusion decoder to generate images that can reveal bit watermarks through the extractor. This approach has become a common practice for subsequent generative model watermarking. AquaLoRA~\cite{feng2024aqualora} referenced Stable Signature's method and shifted the fine-tuning focus to diffusion UNet, improving the precision of watermark decoding. However, directly fine-tuning model weights can lead to a significant decrease in generation quality. LaWa~\cite{rezaei2024lawa} considers using a bit modulation module to modulate bit information between diffusion UNet blocks, training the modulation network so that the extractor can decode bit information. Since the original weights were not modified, the generation quality of the model was maintained. WaDiff~\cite{min2024watermark} further explores generation quality, only modulating the watermark in the last layer of the UNet, to minimize the impact on generation quality.

There are also some training-free approaches including Tree-Ring~\cite{wen2024tree}, RingID~\cite{ci2024ringid}, CRoSS~\cite{yu2024cross} and GaussianShading~\cite{yang2024gaussian} via DDIM Inversion~\cite{mokady2022null}. However, they are only available with diffusion-based models that root watermark in the frequency domain or initial sampling noise, which is kept during the DDIM Inversion process, thus not applicable in 3D Gaussian generation without a diffusion model~\cite{ho2020denoising,song2021denoising,song2021scorebased,li2024omnissr}.

\subsection{Watermarking for 3D Content}
The idea of 3D content watermarking~\cite{yu2003robust} is consistent with the practice of generation model watermarking, which involves leveraging a watermark extractor network first and then fine-tuning the 3D content to decode the bit key correctly~\cite{yoo2022deep,singh2024towards,ohbuchi2002frequency,huang2025geometrysticker,song2024protecting,singh2024towards}. We mainly introduce the watermarking approaches for NeRF~\cite{mildenhall2021nerf} and 3DGS~\cite{kerbl20233d}.

\noindent \textbf{NeRF Watermarking.}
StegaNeRF~\cite{li2023steganerf} is the first exploration into embedding customizable, imperceptible, and recoverable information within NeRF renderings for ownership identification.
The method allows for accurate hidden information extractions from images rendered by NeRF while preserving its visual quality through an optimization framework.
CopyRNeRF~\cite{luo2023copyrnerf} introduces a method to protect the copyright of NeRF models by watermarking, replacing the original color representation in NeRF with a watermarked one. The approach designs a distortion-resistant rendering scheme to ensure robust message extraction in 2D renderings, even when the rendered samples are severely distorted.
WateRF~\cite{jang2024waterf} presents an innovative watermarking method applicable to both implicit and explicit NeRF representations by embedding binary messages during the rendering process.
The method utilizes the discrete wavelet transform in the NeRF space for watermarking and adopts a deferred back-propagation technique along with a patch-wise loss to improve rendering quality and bit accuracy.

\noindent \textbf{3DGS Watermarking.}
GS-Hider~\cite{zhang2024gs} is a steganography framework designed for 3DGS, capable of embedding 3D scenes and images invisibly into original GS point clouds and accurately extracting hidden messages.
The framework replaces the spherical harmonics coefficients of the original 3DGS with a coupled secure feature attribute, using a scene decoder and a message decoder to disentangle the original RGB scene from the hidden message.
GaussianStego~\cite{li2024gaussianstego} is a novel method for embedding images into generated 3DGS assets. It employs an optimization framework that allows for the extraction of hidden information from rendered images while ensuring minimal impact on rendered content quality. 3D-GSW~\cite{jang20243d} introduces a novel watermarking method that embeds binary messages into 3DGS by fine-tuning 3DGS models aiming to protect copyrights.
The method utilizes Discrete Fourier Transform (DFT) to split 3DGS into high-frequency areas, achieving high-capacity and imperceptible watermarking.
GaussianMarker~\cite{huang2024gaussianmarker} introduces an uncertainty-aware digital watermarking method for protecting the copyright of 3DGS models. It proposes embedding invisible watermarks by adding perturbations to 3D Gaussian parameters with high uncertainty, ensuring both invisibility and robustness against various distortions.
\section{Preliminaries for 3DGS Generation}
\label{sec:lgm_representation}
3D Gaussian Splatting (3DGS) is a novel technique in the field of computer graphics and vision that provides an explicit scene representation and enables novel view synthesis without relying on neural networks, unlike NeRF~\cite{kerbl20233d,yu2024mip,huang20242d,charatan2024pixelsplat,lu2024scaffold,chen2025mvsplat}. This method represents scenes using anisotropic 3D Gaussians, which are unstructured spatial distributions. The key to 3DGS is the use of a fast, differentiable GPU-based rendering method to optimize the number, position, and intrinsic properties of the Gaussians, thereby enhancing the quality of scene representation.

The mathematical representation of a 3D Gaussian is given by the formula:
\begin{equation}
    G(\mathbf{x} ; \boldsymbol{\mu}, \mathbf{\Sigma}) = \exp \left(-\frac{1}{2}(\mathbf{x}-\boldsymbol{\mu})^{\top} \mathbf{\Sigma}^{-1}(\mathbf{x}-\boldsymbol{\mu})\right)
\end{equation}
where $\mathbf{x}$ is the point position, the mean $\boldsymbol{\mu}$ and covariance $\mathbf{\Sigma}$ determine spatial distribution. $\mathbf{\Sigma}$ is further calculated via scale matrix $\mathbf{S}$ and rotation matrix $\mathbf{R}$:
\begin{equation}
\label{eq:sigma_calculate}
    \mathbf{\Sigma} = \mathbf{R}\mathbf{S}^{\top}\mathbf{S}\mathbf{R}^{\top}
\end{equation}
Futher, 3D Gaussian can be projected into image space by :
\begin{equation}
    \boldsymbol{\mu}^{'} = \mathbf{PW}\boldsymbol{\mu}, \mathbf{\Sigma}^{'} = \mathbf{JW\Sigma W}^{\top} \mathbf{J}^{\top}
\end{equation}
where $\mathbf{\Sigma}^{'}$ is the covariance matrix in camera space, $\mathbf{W}$ is the viewing transformation matrix, and $\mathbf{J}$ is the Jacobian matrix used to approximate the projective transformation $\mathbf{P}$. For detailed rendering, color $\mathbf{c}$ of specified pixel $\mathbf{p}$ is:
\vspace{-2mm}
\begin{align}
    \label{eq:cpi}
    & \mathbf{c}[\mathbf{p}]=\sum_{i=1}^{N} c_{i} \sigma_{i} \prod_{j=1}^{i-1}\left(1-\sigma_{j}\right) \\
    & \sigma_{i}=\alpha_{i} \exp\left(-\frac{1}{2}(\mathbf{p}-\hat{\boldsymbol{\mu}})^{\top} \hat{\mathbf{\Sigma}}^{-1}(\mathbf{p}-\hat{\boldsymbol{\mu}})\right) \label{eq:sigma_i}
\end{align}
where $N$ denotes the number of sample Gaussian points that overlap pixel $\mathbf{p}$. $c_{i}$ and $\alpha_{i}$ denote the color and opacity of the $i$-th Gaussian, respectively.

Besides the aforementioned representation, it is also possible to present the 3DGS in tensors, which is commonly used in 3D Gaussian generation models~\cite{yi2023gaussiandreamer,tang2025lgm,xu2024grm,he2025gvgen}. Given that one 3DGS can be represented using a 3D-Gaussian distribution, with its mean $\boldsymbol{\mu}$ and variance $\mathbf{\Sigma}$, opacity $\alpha$, scale $\mathbf{S}$, rotation $\mathbf{R}$, and RGB value $\mathbf{c}$\footnote{In standard 3DGS representation, the RGB value is calculated via Spherical Harmonics coefficients. LGM simplifies this by directly storing RGB values in the 3DGS tensor.}, we can embed all these information into one tensor $\mathbf{g}$, as the output target of generation models. In LGM, the output tensor has 14 channels, 3 for point position, 1 for opacity, 3 for scale, 4 for rotation, and 3 for RGB value. The dimension of the tensor under each channel is splatting size $\times$ splatting size.
\section{Method}
\subsection{Task Settings \& Potential Intuitive Approaches}
\label{sec:intuitive}
\noindent \textbf{Task Settings.} Our watermarking framework is designed to protect the copyright of 3DGS generative models and their outputs. The copyright owner of a 3DGS generative model specifies a bit string as the copyright identifier, which must be accurately decoded from the generated results. Since 3DGS structures are typically viewed after rendering as images, we choose to decode the bit copyright identifier from the rendered outputs. In this context, 3DGS models are publicly uploaded to the internet, necessitating a certain level of robustness against common attack types, such as point cloud pruning, and augmentation attacks to rendered images.

\noindent \textbf{Intuitive Approaches.} Considering that our work is the first to embed bit watermark in 3DGS generative models, we present some potential alternative approaches to illustrate the effectiveness of our method, both \textit{post-generation} and \textit{in-generation} watermarking methods, discussing why they are not suitable for watermarking 3DGS generative models. Performance comparisons are detailed in the Sec.~\ref{sec:comparison}.

\noindent \ding{113}~\textbf{3DGS+HiDDeN}: This approach involves training an additional watermark decoder to extract bit information from the rendered outputs of the 3DGS. We choose the HiDDeN~\cite{zhu2018hidden} network, a commonly used architecture for bit addition and decoding, to facilitate bit extraction. However, while this method might be effective in image watermarking, it \textbf{struggles to decode correctly} on rendered results of 3DGS, as rendering is a \textbf{strong degradation} for 3DGS.

\noindent \ding{113}~\textbf{3DGS+WateRF}: It is also possible to embed watermark into 3DGS objects using the methods applied in WateRF~\cite{jang2024waterf}, which is proposed to hide watermark into NeRF. However, due to the different natures of 3DGS and NeRF, approaches available for NeRF watermarking might lead to \textbf{inferior performance} for 3DGS watermarking.

\noindent \ding{113}~\textbf{Fine-tuning UNet}: Besides adding modulation modules to the fixed generation model, another possible idea is to directly fine-tune the weights of the generation model, making the model output the 3DGS while also decoding bits correctly. This approach works well in diffusion-based image generation models as its UNet is trained through a denoising process, making latent code robust to noise, thus minor weight changes do not significantly affect the generation quality. In contrast, the latent codes in the UNet of LGM are more fragile, thus weight modifications from fine-tuning can drastically alter the attributes of the generated 3DGS point clouds, thereby \textbf{hard to converge} and \textbf{impacting the generation quality}.

\noindent \ding{113}~\textbf{Extract from 3DGS directly}: As discussed in Sec.~\ref{sec:lgm_representation}, the 3DGS contains dedicated channels representing RGB values, which can be reshaped to match the image size for bit decoding. However, the visualization results of the RGB values in the 3DGS tensor show a substantial \textbf{domain gap from natural images}, making it challenging for a pre-trained bit decoder to extract the embedded information accurately.
\subsection{Overview}
\begin{figure*}
    \centering
    \includegraphics[width=1\linewidth]{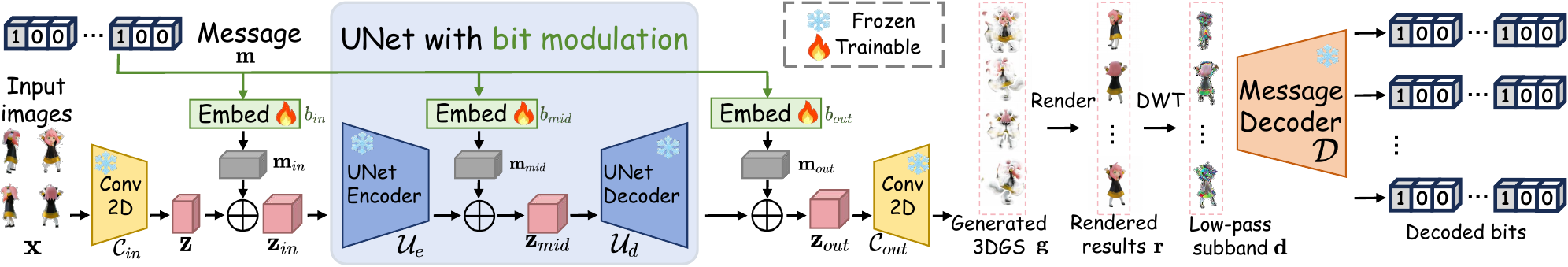}
    \caption{An overview architecture of our proposed 3D generation model watermarking framework, with a detailed illustration of bit modulation embedding and bit decoding process. The input images are transformed into features and processed by the UNet, whose output is converted to 3DGS tensor. The input bit is modulated, multiplied by adaptive coefficients, and embedded at the input, middle, and output of the UNet. This watermark can be extracted from the 3DGS rendering results after DWT.}
    \label{fig:overview}
    \vspace{-2mm}
\end{figure*}
% 先用中文写个粗糙版本
% 第一段：框架，输入，输出
% 第二段：输入处理，
% 第三段：输入图像，unet结构介绍
% 第四段：bit embedding，包括网络概要、自适应系数（采用现在的就可以）
% 第五段：结果渲染
% 第六段：结果解码
Our proposed 3DGS generative model watermarking framework is based on the current commonly used 3DGS generation state-of-the-art method LGM~\cite{tang2025lgm}. By inputting an image $\mathbf{I}$ and a user-specified binary string $m$ as the watermark, our framework will output a high-quality 3DGS object visually consistent with the input image, and the watermark can be decoded in rendered results $\mathbf{R}$.
For processing input data, we first use a pre-trained MVDream~\cite{shi2023mvdream} pipeline to generate 4 multi-view images $\mathbf{x}$ of input image $\mathbf{I}$ to provide detailed multi-view visual information, which is further transformed as features via Conv2D layer $\mathcal{C}_{in}$. The input watermark $m$ is transformed into binary tensor $\mathbf{m}$. The images feature $\mathbf{z}$ are then sent into generation UNet $\mathcal{U}$. The UNet has 5 encoder blocks $\mathcal{U}_{e}$ , and 3 decoder blocks (with 1 middle block) $\mathcal{U}_{d}$. 

To effectively embed bits into generation UNet, we implement an adaptive bit modulation module, denoted as $\mathcal{B}$, allowing for seamless integration of the watermark within the generative process. The modulation modules are located separately at the image input $\operatorname{b}_{in}$, the output of the encoder blocks $\operatorname{b}_{mid}$, and the output of UNet $\operatorname{b}_{out}$. We observe that the value range of 3DGS is consistent with the binary 0-1 values of bits, so directly combining the bit tensor with the block output would affect the generation quality. Therefore, we designed learnable adaptive coefficients $\alpha, \beta \; \text{and} \; \gamma$ for each bit embeddings to avoid this issue. A detailed modulation process is shown in Algo.~\ref{alg:watermark}.

The tensor output by the LGM, denoted as $\mathbf{z}_{out}$, is transformed into the shape of the 3DGS tensor (mentioned in Sec.~\ref{sec:lgm_representation}) via a Conv2D layer $\mathcal{C}_{out}$, denoted as $\mathbf{g}$. It is further rendered into images of 180 different views $\mathbf{r}$. From these rendered images, we utilize the output low-pass subband of Discrete Wavelet Transform (DWT)~\cite{akansu1990perfect}, denoted as $\mathbf{d}$, to extract the bit results. For the decoding network, we draw inspiration from the task setup utilized in Stable Signature~\cite{fernandez2023stable}, employing a pre-trained watermark extractor network $\mathcal{D}(\cdot)$ from HiDDeN~\cite{zhu2018hidden} to enhance the decoding accuracy and efficiency. An overall illustration of our proposed framework is shown in Fig.~\ref{fig:overview}.

\subsection{Adaptive Bit Modulation}
\label{sec:modulation}
The bit message input by the user is a 0-1 binary string, and we first convert it into a binary tensor $\mathbf{m}$. Next, we employ bit modulation module $\mathcal{B}$ to embed this tensor into outputs of various blocks in the UNet $\mathcal{U}$, and weights of embedding networks are initialized as 0. Compared to directly fine-tuning the UNet, this approach significantly reduces GPU memory consumption, as we only need to train specific modulation layers. Additionally, the difference between the UNet inference results before and after modulation is minimal via such approach, which effectively preserves the generative quality of the 3DGS outputs.

Given the large range of possible values for the 0-1 bits, which is close to the value range of the tensors output by the LGM, directly embedding this binary tensor into the intermediate results of the UNet could lead to a degradation in quality. Furthermore, the impact of different blocks within the UNet on the generated results varies; thus, directly embedding the 0-1 bit tensor into the intermediate results is not a reasonable approach. Based on this analysis and observation, we utilize a set of learnable adaptive coefficients $\alpha, \beta, \gamma$ to constrain the degree of modulation.

\begin{figure}
    \centering
    \includegraphics[width=1\linewidth]{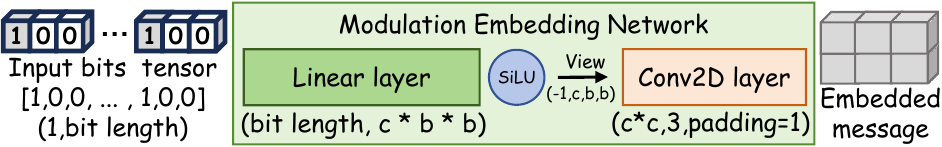}
    \caption{Illustration of bit embedding network. ``c" denotes channel size, and ``b" denotes block size.}
    \label{fig:net}
\vspace{-4mm}
\end{figure}

The specific network architecture is structured as follows: the bit modulation modules $\mathcal{B}$ are located at three stages within the UNet of the LGM, specifically (1) in the input image, named $\operatorname{b}_{in}$, (2) in the middle of UNet blocks  $\operatorname{b}_{mid}$, and (3) in the output of UNet $\operatorname{b}_{out}$. 
The embedding network for the input 0-1 bits consists of a linear layer followed by a SiLU~\cite{silu2018} activation function, ultimately producing an output of the appropriate size. This result is then processed through a 2D convolutional layer, multiplied by an adaptive coefficient denoted as $\alpha$, and added to the input image of the UNet. This modulation process is formulated as follows:
\begin{equation}
    \mathbf{m}_{in} = \alpha \operatorname{b}_{in}(\mathbf{m}), \;
    \mathbf{z}_{in} = \mathbf{z} + \mathbf{m}_{in}
\label{eq:input_modulation}
\end{equation}
For the modulation modules positioned in the output of UNet, the process is similar. In both cases, the 0-1 bits are embedded into tensors using a linear layer, the SiLU activation function, and 2D convolutional layers to extract further features. The resulting tensor is then multiplied by adaptive coefficients, referred to as $\beta$ and $\gamma$, and added to the intermediate results of the corresponding blocks.
The processes are formulated as:
% \begin{equation}
% \begin{aligned}
%     & \mathbf{m}_{down} = \mathbf{b}_{down}(\mathbf{m}) \\
%     & \mathbf{x} = \mathbf{x} + \beta \; \mathbf{m}_{down} \\
%     & \mathbf{x} = \mathbf{u}_{up}(\mathbf{x})
% \end{aligned}
% \label{eq:downmid_modulation}
% \end{equation}
% and 
% \begin{equation}
% \begin{aligned}
%     & \mathbf{m}_{up} = \mathbf{b}_{up}(\mathbf{m}) \\
%     & \mathbf{x} = \mathbf{x} + \gamma \; \mathbf{m}_{up}
% \end{aligned}
% \label{eq:up_modulation}
% \end{equation}
\vspace{-1mm}
\begin{align}
    \label{eq:downmid_modulation}
    & \mathbf{m}_{mid} = \beta \operatorname{b}_{mid}(\mathbf{m}), \;
    \mathbf{z}_{mid} = \mathcal{U}_{e}(\mathbf{z}_{in}) + \mathbf{m}_{mid} \\ 
    & \mathbf{m}_{out} = \gamma \operatorname{b}_{out}(\mathbf{m}), \;
    \mathbf{z}_{out} = \mathcal{U}_{d}(\mathbf{z}_{mid}) + \mathbf{m}_{out}  \label{eq:up_modulation}
\end{align}
% \vspace{-1mm}
% \begin{equation}
%     \mathbf{m}_{mid} = \beta \operatorname{b}_{mid}(\mathbf{m}), \;
%     \mathbf{z}_{mid} = \mathcal{U}_{e}(\mathbf{z}_{in}) + \mathbf{m}_{mid}
% \label{eq:downmid_modulation}
% \end{equation}
% \begin{equation}
%     \mathbf{m}_{out} = \gamma \operatorname{b}_{out}(\mathbf{m}), \;
%     \mathbf{z}_{out} = \mathcal{U}_{d}(\mathbf{z}_{mid}) + \mathbf{m}_{out}
% \label{eq:up_modulation}
% \end{equation}
This structured approach ensures effective integration of the watermarking information while maintaining the overall generative quality of the model.
The specific network module is shown in Fig.~\ref{fig:net}.
\begin{algorithm}
    \scriptsize
    \caption{GaussianSeal training process}
    \label{alg:watermark}
    \KwIn{Input multi-view images $\mathbf{x}$, secret message $ \mathbf{m}$, Conv2D layers $\mathcal{C}_{in}$ and $\mathcal{C}_{out}$, UNet $\mathcal{U}=\{\mathcal{U}_{e}, \mathcal{U}_{d} \}$, bit modulation blocks $\mathcal{B}=\{\operatorname{b}_{in}, \operatorname{b}_{mid}, \operatorname{b}_{out} \}$, adaptive coefficients $\alpha, \beta, \gamma$, watermark decoder $\mathcal{D}$}
    \KwOut{Bit modulation blocks and adaptive coefficients}
    \For{$\text{epoch}=1$ \KwTo $\text{total epochs}$}{
        $\mathbf{z}=\mathcal{C}_{in}(\mathbf{x})$ \\
        Get $\mathbf{z}_{in}$ via bit embedding as Eq.~(\ref{eq:input_modulation}). \\
        
        Get $\mathbf{z}_{mid}$ via bit embedding as Eq.~(\ref{eq:downmid_modulation}). \\
    
        Get $\mathbf{z}_{up}$ via bit embedding as Eq.~(\ref{eq:up_modulation}). \\
        $\mathbf{g} = \mathcal{C}_{out}(\mathbf{z}_{up})$
    
        $\mathbf{r} = \text{3DGS Render}(\mathbf{z}_{up})$ \\
        $\mathbf{d} = \text{DWT}(\mathbf{r})$ \\

        Calculate loss via Eq.~(\ref{eq:overall}). \\
        $\mathcal{B} = \mathcal{B}\text{.update\(\)}$ \\
        $\alpha, \beta, \gamma = \alpha\text{.update\(\)}, \beta\text{.update\(\)}, \gamma\text{.update\(\)}$
        
    }    
    \Return Bit modulation blocks $\mathcal{B}$, adaptive coefficients $\alpha, \beta, \gamma$
            % \end{algorithmic}
    
\end{algorithm}
\vspace{-4pt}
% ALGO
\subsection{Bit Message Decoding}
After the generation model gets embedded with the bits, we anticipate that they can be accurately decoded to validate the copyright of both the model and the generated outputs. To achieve this, we explored several common decoding targets, including the 3DGS tensor generated, and the rendered images of the 3DGS.

In our investigations, we focus on the bit extraction performance of each target. We align our approach with the established watermarking methodology, emphasizing robustness and fidelity, which led us to choose the DWT-transformed outputs of the 3DGS renderings as the primary source for decoding the bit results. The DWT is particularly effective for capturing the essential features of the rendered images while preserving the watermark information, and this claim is supported in ablation study in Sec.~\ref{sec:ablation_studies}. For the watermark extracting network, we leverage HiDDeN~\cite{zhu2018hidden}'s watermark decoder and pre-train this network on Objaverse~\cite{objaverse} corresponding dataset, as there is a severe domain gap between datasets used in pre-training of HiDDeN and Objaverse.
\begin{figure*}
    \centering
    \includegraphics[width=1\linewidth]{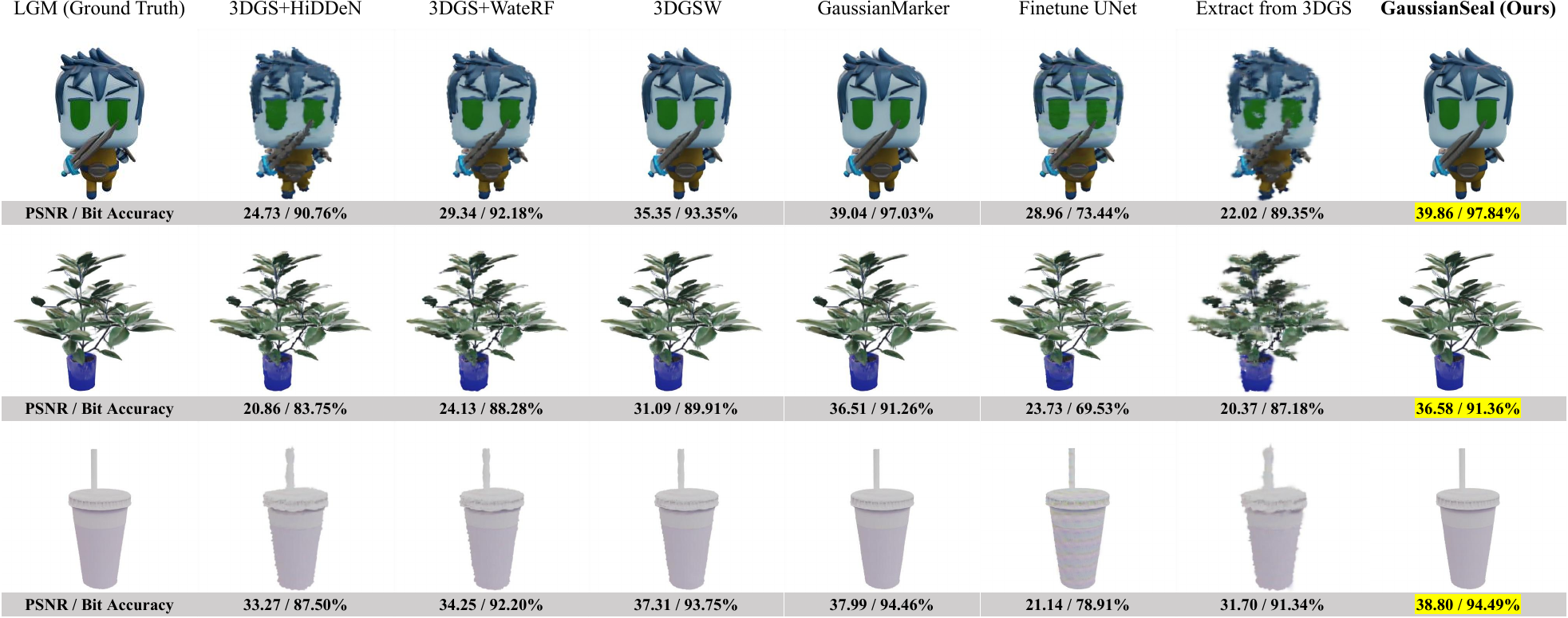}
    \caption{Visualized results of rendered 3D objects watermarked via compared methods, and our GaussianSeal. These results are on 16 bits.}
    \label{fig:compare}
    \vspace{-4mm}
\end{figure*}

\subsection{Training Details}

Our training approach employs an end-to-end strategy, where the loss function comprises two components: the bit information loss and the consistency loss between the 3D generation results before and after modulation and the rendered outputs.
For the bit loss, we utilize binary cross-entropy (BCE) loss to effectively capture the discrepancies between ground-truth message and decoded result from rendered images, as Eq.~(\ref{eq:loss_msg}).
\begin{equation}
\label{eq:loss_msg}
    \mathcal{L}_{msg} = \text{BCE}(\mathbf{m}, \mathcal{D}(\mathbf{d}))
\end{equation}
Regarding the consistency loss of the generated results, we simultaneously consider the consistency of the tensor and the consistency of the rendered outputs using mean-squared error (MSE) loss, as Eq.~(\ref{eq:loss_con}):
\begin{equation}
\label{eq:loss_con}
    \mathcal{L}_{gs} = \text{MSE}(\mathbf{g}, \mathbf{g}_{clean}), \
    \mathcal{L}_{rgb} = \text{MSE}(\mathbf{R}, \mathbf{R}_{clean})
\end{equation}
where $\mathbf{g}_{clean}$ and $\mathbf{R}_{clean}$ denote 3DGS tensor and rendered results from original model.
Overall training loss is as follows:
\begin{equation}
\label{eq:overall}
    \mathcal{L} = \mathcal{L}_{msg}
                + \lambda_{gs} \; \mathcal{L}_{gs}
                + \lambda_{rgb} \; \mathcal{L}_{rgb}
\end{equation}
The weights among the various loss components, $\lambda_{gs}$ and $\lambda_{rgb}$, are determined through a grid search process, ensuring optimal performance. Additionally, the adaptive modulation coefficients discussed in Sec.~\ref{sec:modulation} are updated using an auxiliary optimizer, allowing for fine-tuning of the modulation process and enhancing the overall effectiveness of the watermarking framework. The training process of GaussianSeal with detailed modulation is shown in Algo.~\ref{alg:watermark}, with parameters of bit modulation network and adaptive coefficients trained through this pipeline.
% \begin{table*}[h!]
% \centering
% \begin{tabular}{|c|c|c|c|c|c|c|c|c|c|c|c|}
% \hline
% \multirow{3}{*}{AAA} & \multirow{3}{*}{BBB} & \multicolumn{10}{c|}{CCC} \\ \cline{3-12}
%                      &                      & Data1 & Data2 & Data3 & Data4 & Data5 & Data6 & Data7 & Data8 & Data9 & Data10 \\ \hline
% Data11               & \multirow{3}{*}{\shortstack{Data12 \\ Data13 \\ Data14}} & Data15 & Data16 & Data17 & Data18 & Data19 & Data20 & Data21 & Data22 & Data23 & Data24 \\ \cline{3-12}
%                      &                      & Data25 & Data26 & Data27 & Data28 & Data29 & Data30 & Data31 & Data32 & Data33 & Data34 \\ \hline
% \end{tabular}
% \caption{1}
% \end{table*}
\begin{table*}
    \centering
    \begin{tabular}{cc|ccc|c|c}
    \toprule[1.5pt]
       Method Type  &  Method &  PSNR$\uparrow$ & SSIM$\uparrow$ & LPIPS$\downarrow$ & Bit Accuracy$\uparrow$ & Time (s)$\downarrow$\\
       \hline
       \multirow{4}{*}{Post-Generation}  & 3DGS+HiDDeN~\cite{zhu2018hidden} & 31.8991 & 0.9067 & 0.0082 & 92.83\% & 786.40\\
       % \hline
       & 3DGS+WateRF~\cite{jang2024waterf} & 32.1800 & 0.9486 & 0.0084 & 94.05\% & 2184.25\\
       % \hline
       & 3D-GSW~\cite{jang20243d}  & 33.5937 & 0.9485 & \boldblue{0.0071} & 94.38\% & 475.23\\
       & GaussianMarker~\cite{huang2024gaussianmarker}  & \boldblue{37.4583} & \boldblue{0.9813} & 0.0075 & \boldblue{97.19\%} & 1228.71\\
       \hline
       \multirow{3}{*}{In-Generation}  & Fine-tune UNet &  21.5227 & 0.9279 & 0.0819 & 64.06\% & \boldblue{0.18}\\
       % \hline
       & Extract from 3DGS  & 22.6368 & 0.9525 & 0.0096 & 89.58\% & \boldred{0.12}\\
       % \hline
       & GaussianSeal (Ours)  & \boldred{38.0228} & \boldred{0.9892} & \boldred{0.0034} & \boldred{97.93\%} & \boldblue{0.18}\\
       \bottomrule[1.5pt]
    \end{tabular}
    \vspace{-2mm}
    \caption{Quantitative results of comparison between our proposed framework and other approaches. These results are on 16 bits. Best results are shown in \boldred{red}, and second best results are shown in \boldblue{blue}.}
    \label{tab:main}
    \vspace{-1mm}
\end{table*}
% Notably, our experimental validation reveals that the updates to RGB, scale, and opacity within the LGM's representation of the 3DGS significantly impact the quality of the generated 3DGS. In contrast, the rotation matrix exhibits a negligible effect on the rendering results of the 3DGS. 
% This finding is substantiated in Sec.~\ref{sec:attributes} of our experiments.
\begin{table*}
    \centering
    \begin{tabular}{cc|ccc|c|c}
    \toprule[1.5pt]
       Method Type  &  Method &  PSNR$\uparrow$ & SSIM$\uparrow$ & LPIPS$\downarrow$ & Bit Accuracy$\uparrow$ & Time (s)$\downarrow$\\
       \hline
       \multirow{3}{*}{Post-Generation}  
       & 3DGS+WateRF~\cite{jang2024waterf} & 32.8751 & 0.9446 & 0.0088 & 89.06\% & 2259.42\\
       % \hline
       & 3D-GSW~\cite{jang20243d}  & 33.1479 & 0.9539 & 0.0079 & 92.63\% & \boldblue{763.54}\\
       & GaussianMarker~\cite{huang2024gaussianmarker}  & \boldblue{35.6208} & \boldblue{0.9541} & \boldblue{0.0056} & \boldblue{94.71\%} & 1292.95\\
       \hline
       \multirow{1}{*}{In-Generation} 
       & GaussianSeal (Ours)  & \boldred{36.8051} & \boldred{0.9583} & \boldred{0.0045} & \boldred{96.58\%} & \boldred{0.18}\\
       \bottomrule[1.5pt]
    \end{tabular}
    \vspace{-2mm}
    \caption{Comparison of bit embedding performance in 32 bits. Best results are shown in \boldred{red}, and the second best results are shown in \boldblue{blue}.}
    \label{tab:32bit}
    \vspace{-2mm}
\end{table*}
\section{Experiments}

\subsection{Experimental Settings}
\noindent \textbf{Datasets and Pre-training.}
We select Objaverse dataset~\cite{objaverse} as our training and validation dataset, from which we randomly sample 10000 objects for training and 100 objects for validation. Specifically, the data for each object consists of images captured from 38 different viewpoints, along with the corresponding camera intrinsic and extrinsic parameters, viewpoint information, and rotation matrices. All images are standardized to the size of 512$\times$512.
\begin{table}
    \centering
    \resizebox{1.\linewidth}{!}{
    \begin{tabular}{c|ccc}
    \toprule[1.5pt] % \rowcolor{color3} Decoding Target
        Decoding Target & PSNR$\uparrow$ & SSIM$\uparrow$ & Bit Accuracy$\uparrow$ \\
        \hline
        3DGS Tensor & 22.6368 & 0.9525 & 89.58\%\\
        \hline
        Rendered Images & 34.2383 & 0.9675 & 96.09\%\\
        \hline
        Rendered Images + DWT & \boldred{38.0228} & \boldred{0.9892} & \boldred{97.93\%}\\
    \bottomrule[1.5pt]
    \end{tabular}
    }
    % \vspace{-2mm}
    \caption{Results of performance comparison among different decoding targets. Best results are shown in \boldred{red}.}
    \label{tab:decoding_targets}
    \vspace{-4mm}
\end{table}
% For the decoding phase, we utilized the HiDDeN~\cite{zhu2018hidden} network as referenced in Stable Signature, employing the same pre-trained weights to ensure consistency and enhance the performance of our decoding process. This choice allows us to leverage established methodologies while adapting them to our specific watermarking framework for 3DGS generation.
For the 3DGS generation model, we use pre-trained weights of LGM~\cite{tang2025lgm} without fine-tuning or modification. For the watermark decoder, we pre-train HiDDeN~\cite{zhu2018hidden} decoder
for 16 and 32 bits and fix the parameters during the training process.

\noindent \textbf{Metrics.}
Our framework aims to achieve higher accuracy in bit decoding while minimizing the impact on the original 3D generative model, thus we evaluate the performance of our proposed watermarking framework in \textbf{accuracy} and \textbf{visual consistency}.
The bit accuracy process is calculated via code from ~\cite{wen2024tree}. For visual consistency, the metrics include PSNR, SSIM, and LPIPS~\cite{zhang2018lpips} between the rendered results of the watermarked model and the original model. We also compare the watermarking time for each 3DGS object, which \textit{in-generation} methods have significant advantages over \textit{post-generation} methods.

\noindent \textbf{Implementation Details.}
For training bit modulation modules, we choose AdamW~\cite{loshchilov2019decoupled} as the optimizer with learning rate 1e-4. For adaptive coefficients $\alpha$, $\beta$, and $\gamma$, we initialize them to 0.1 and let another AdamW optimizer with a learning rate 1e-3 to update them. For loss balancing weights $\gamma_{gs}$ and $\gamma_{rgb}$, we choose $\gamma_{gs}=1000$ and $\gamma_{rgb}=300$. Detailed ablation for these two hyper-parameters is shown in Sec.~\ref{sec:lossweight}. The bit message length is set as 16 and 32 in evaluation (Sec.~\ref{sec:comparison}), and set as 16 in other experiments.
All experiments are done on NVIDIA GTX 3090Ti GPU, and the batch size is set as 2.
\subsection{Evaluation of proposed GaussianSeal}
\label{sec:comparison}
\noindent \textbf{Baselines.} As we are the first to propose a bit watermarking method for 3D generative models, we select comparison methods to demonstrate the effectiveness of our approach including \textit{post-generation} 3D object watermarking and \textit{in-generation} intuitive watermarking methods mentioned in Sec.~\ref{sec:intuitive}, including:
(1) \textbf{3DGS+HiDDeN}~\cite{zhu2018hidden}: train a watermark decoder to extract bits correctly; (2) \textbf{3DGS+WateRF}~\cite{jang2024waterf}: train a watermarked 3DGS object via WateRF approach; (3) \textbf{3D-GSW}~\cite{jang20243d} current 3DGS watermarking state-of-the-art method via fine-tuning 3DGS representation; (4) \textbf{GausssianMarker}~\cite{huang2024gaussianmarker} current state-of-the-art 3DGS watermarking method via uncertainty estimation; (5) \textbf{Fine-tune UNet}: fine-tuning generation UNet to decode bits correctly; (6) \textbf{Extract from 3DGS}: use our bit modulation module, bit extract the watermark directly from the tensor representation of the 3DGS.
% For post-generation 3D object watermarking, we choose watermark method for single 3DGS objects, including , and NeRF watermarking methods, including  and , mentioned in Sec.~\ref{sec:intuitive}. Second, we compare the potential intuitive approaches from other model watermarking methods applied to the 3D generative models, including (1)  instead of adding bit modulation module, and (2) , instead of from the rendered result after DWT transformation. We compare their performance in bit accuracy and visual consistency. 

\begin{figure}
    \centering
    \includegraphics[width=1\linewidth]{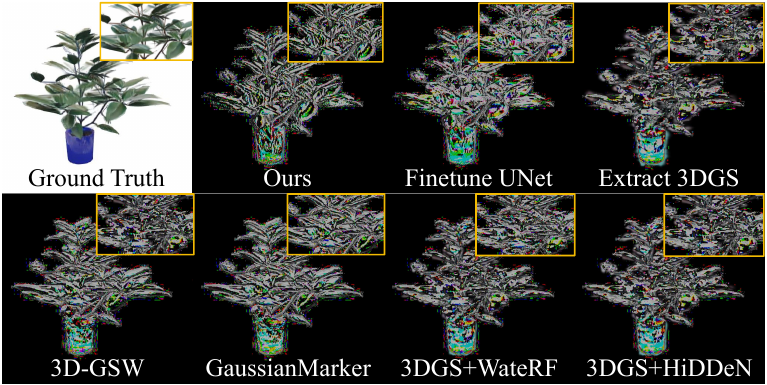}
    \caption{Residual for watermarked rendered images and originally generated rendered results. We put a detailed view in the center of the example plant image. It shows that our watermark keeps more detail of the original image, with the shape of branches and leaves unchanged. Other watermark methods make details broken, resulting in worse quality.}
    \label{fig:residual_image}
    \vspace{-4mm}
\end{figure}

Quantitative results are shown in Tab.~\ref{tab:main}, and the visual experimental results are presented in Fig.~\ref{fig:compare}. From the experimental results, we observe that our watermarking framework achieves the best bit decoding accuracy and visual consistency.
We also provide residual images for watermarked rendered images and vanilla-generated rendered results in Fig.~\ref{fig:residual_image}. It shows that our watermark preserves the detailed structure of generated images.
\begin{table*}[h!]
\centering

\begin{tabular}{c|cccccc}
\toprule[1.5pt]
\multirow{3}{*}{Method / Attack} & \multicolumn{6}{c}{Bit Accuracy(\%)$\uparrow$} \\
% \cline{2-7}
& Gaussian Noise & Crop & Rotation & Brightness & JPEG Compression & Gaussian Blur  \\
% \cline{2-7}
& ($\mu$ = 0.1) & (40\%) & ($\text{60}^{\circ}$) & (2.0) & (Quality 10\%) & (kernel size = 5)  \\
\hline
3DGS+WateRF~\cite{jang2024waterf} & 95.50 & \boldblue{95.73} & 92.59 & 92.73 & 95.42 & 95.20\\

3D-GSW~\cite{jang20243d} & 94.93 & \boldred{95.75} & 95.25 & \boldblue{94.84} & \boldblue{96.12} & 96.29\\
GaussianMarker~\cite{huang2024gaussianmarker} & \boldblue{96.18} & 95.37 & 93.94 & 94.37 & 90.63 & \boldred{98.36}\\
\hline
GaussianSeal (Ours) & \boldred{96.32} & 95.20 & \boldred{96.34} & \boldred{97.26} & \boldred{96.81} & \boldblue{97.02}\\
\bottomrule[1.5pt]
\end{tabular}
\vspace{-2mm}
\caption{Robustness test result on common image augmentation attacks. Best results are shown in \boldred{red}, and second best results shown in \boldblue{blue}.}
\label{tab:robust_augment}
\vspace{-4mm}
\end{table*}

\subsection{Ablation Study}
\label{sec:ablation_studies}
Here we discuss the key technical choices in our method.
\begin{table}
    \centering
    \resizebox{0.95\linewidth}{!}{
    \begin{tabular}{ccc|cc}
    \toprule[1.5pt]
        Input & Middle of blocks & Output & PSNR & Bit Acc\\
        \hline
        $\times$ & $\times$ & $\times$ & $\inf$ & 64.53\%\\
        $\checkmark$ & $\times$ & $\times$ & 38.7487 & 89.20\%\\
        $\times$ & $\checkmark$ & $\times$ & 38.7396 & 84.31\% \\
        $\times$ & $\times$ & $\checkmark$ & 38.7589 & 92.96\% \\
        $\checkmark$ & $\checkmark$ & $\times$ & 38.1323 & 97.65\%\\
        $\checkmark$ & $\times$ & $\checkmark$ & 38.1967 & 93.19\%\\
        $\times$ & $\checkmark$ & $\checkmark$ & 38.2432 & 95.30\%\\
        $\checkmark$ & $\checkmark$ & $\checkmark$ & 38.0228 & 97.93\%\\
        \bottomrule[1.5pt]
    \end{tabular}
    }
    % \vspace{-2mm}
    \caption{Ablation of bit modulation on different positions in UNet.}
    \label{tab:positions}
    \vspace{-2mm}
\end{table}
\noindent \textbf{Selection of Watermark Decoding Target.}
The choice of decoding target for watermarking is non-trivial. Unlike image generation models that directly decode bits from generated results, we opt to decode from rendered results via DWT. This decision is guided by both the practical insights from WateRF~\cite{jang2024waterf} and our own experimental validation. Results are shown in Tab.~\ref{tab:decoding_targets}, which shows that the DWT sub-band of rendered results is easier to get decoded bits.

\noindent \textbf{Discussion of Modulation Modules.}
We incorporate modulation modules at three locations: in the input image feature, in the middle of UNet blocks, and in the output of UNet. What if we add watermarking to only some of these positions? We test with different configurations and concluded that including bit modulation modules at all three locations improves bit decoding performance. Results are presented in Tab.~\ref{tab:positions}. It is shown that all modulation modules function effectively, contributing to precise bit decoding. Without significantly impacting the quality of generation, we opt to add modulation modules at all three locations. 

\begin{table}
    \centering
    \begin{tabular}{cc|cc}
    \toprule[1.5pt]
       $\lambda_{gs}$  & $\lambda_{rgb}$ & PSNR & Bit Acc \\
       \hline
       3000  & 100 & 43.1517 & 86.71\% \\
       2000  & 100 & 40.1112 & 91.40\%  \\
       1000  & 100 & 35.0509 & 92.18\% \\
       1000  & 200 & 35.2704 & 95.31\% \\
       \textbf{1000}  & \textbf{300} & \textbf{38.0228} & \textbf{97.93\%} \\
       1000  & 400 & 38.1557 & 96.09\% \\
       500   & 100 & 32.6613 & 70.31\% \\
    \bottomrule[1.5pt]
    \end{tabular}
    % \vspace{-2mm}
    \caption{Ablation results of different choices on weights of Gaussian tensor loss and RGB loss. Our choice is shown in \textbf{bold}.}
    \label{tab:lossweight}
    \vspace{-2mm}
\end{table}
\begin{table}
    \centering
    \begin{tabular}{c|cccc}
    \toprule[1.5pt]
        Ratio & PSNR & SSIM & LPIPS & Bit Acc\\
        \hline
        5\% & 27.6489 & 0.9762 & 0.0082 & 96.37\%\\
        10\% & 25.3123 & 0.9156 & 0.0097 & 92.75\%\\
        15\% & 24.3196 & 0.8978 & 0.0183 & 91.96\%\\
        25\% & 23.0801 & 0.8803 & 0.0245 & 89.20\%\\
    \bottomrule[1.5pt]
    \end{tabular}
    \vspace{-2mm}
    \caption{Robustness analysis of proposed watermarking framework under different 3DGS pruning ratios.}
    \label{tab:robust}
    \vspace{-6mm}
\end{table}

\noindent \textbf{Balancing Loss Weights.}
\label{sec:lossweight}
We fix the bit message loss weight at 1 and performed a grid search for the Gaussian tensor and RGB rendering loss weights. The results of this grid-search process are shown in Tab.~\ref{tab:lossweight}. To balance the generation quality (measured by PSNR) and bit decoding accuracy, we choose $\lambda_{gs}$ as 1000 and $\lambda_{rgb}$ as 300.

\subsection{Method Analysis}
\begin{figure}
    \centering
    \includegraphics[width=1\linewidth]{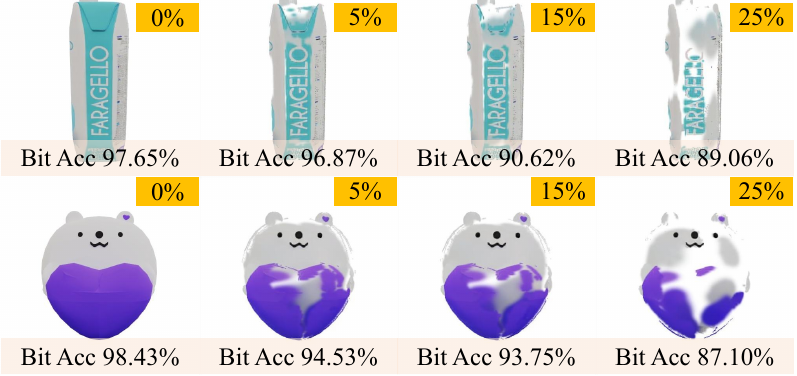}
    \caption{Visualization of pruned 3DGS point clouds, with each decoded accuracy. The ratio on the image denotes the prune ratio.}
    \label{fig:prune}
    \vspace{-4mm}
\end{figure}

\noindent \textbf{Robustness of Our Watermark.}
Current attacks on 3DGS objects primarily involve pruning, where the Gaussian structure is disrupted by reducing the number of Gaussian points in the point cloud. To assess the robustness of our watermarking framework against this type of attack, we conducted tests with results shown in Tab.~\ref{tab:robust}, and a visualized result of pruned 3DGS point cloud with decoded accuracy is shown in Fig.~\ref{fig:prune}.
We also test our watermark framework's robustness on common image augmentation attacks. Results are shown in Tab.~\ref{tab:robust_augment}.
Results demonstrate that our watermark is robust to common attacks, both in 3D and 2D domain.

\noindent \textbf{Security Analysis of Our Watermark.}
Additionally, we evaluate the exposure risk of our embedded watermark using the open-source tool StegExpose~\cite{boehm2014stegexpose}. Detailed results are shown in Supplementary Materials, indicating that our watermark is sufficiently secure and difficult to detect. 

% \noindent \textbf{Robustness of Gaussian Properties to Watermark Modulation}
% \label{sec:attributes}
% As previously mentioned, the output of the LGM~\cite{tang2025lgm} consists of 14 channels representing the Gaussian properties, including the mean of the 3D Gaussian distribution, opacity, scale, rotations, and the RGB values of the point. In addition to the adaptive modulation approach, we also test watermarking on individual attributes to evaluate the robustness of different attributes to watermark modulation. Results are provided in Supplementary Materials.
% 改下说辞?
\subsection{Limitation and Discussion}
The common issue in the current field of model watermarking is the large demand for dataset scale. T2I model watermarking methods such as LaWa~\cite{rezaei2024lawa} and AquaLoRA~\cite{feng2024aqualora}, as well as our GaussianSeal for 3D generative model, all require large amounts of data for training to simultaneously achieve convergence of the watermark loss and maintain generation quality.
Our watermarking technique is also limited to protecting the copyright of 3DGS objects generated by AI models. In future work, we aim to expand our approach to include watermarking for point clouds and meshes rendered from generated 3DGS objects. This will enable us to achieve comprehensive watermark coverage across the entire process of 3D generation, thereby enhancing the robustness and versatility of our watermarking framework.
\vspace{-2mm}
\section{Conclusion}
In this paper we propose the first bit watermarking framework for 3D Gaussian generative models, named GaussianSeal, to protect the copyright of 3D generative models and their generated results. We design an adaptive bit watermark modulation module that is able to effectively embed bit watermarks into generative model's blocks, achieving accurate watermark decoding while maintaining the consistency and fidelity of original generation quality for both 3D Gaussian model and the rendered images, which is challenging. In future work, we aim to extend GaussianSeal to generalize across multiple bit messages, enabling both model owner verification and generator tracing. We will also consider expanding the decoding targets of the watermark to rendered results such as point clouds and meshes from 3DGS, thereby achieving versatile watermarking across multiple modalities. In summary, our method effectively protects the copyright of AIGC models and corresponding 3D assets. 
% \looseness=-1

\clearpage
\maketitlesupplementary

\noindent In the supplementary materials, we demonstrate additional experimental results, implementation details, discussion, and analysis as follows.

\tableofcontents
\newpage

\section{Implementation Details}
\label{sec: details}

\subsection{Dataset Details}
\label{sec: dataset_details}
For the Objaverse dataset, each object has 38 views, and each view is 512$\times$512 with an RGB value range from 0 to 1 and set in white background. The camera system loaded in raw dataset is Blender world and OpenCV camera, and we transform it into OpenGL world and OpenGL camera.

For the NeRF Synthetic dataset~\cite{mildenhall2020nerf} in additional evaluation, for each object we choose one view that contains most of visual information of the object, and follow the same LGM inference pipeline used in the evaluation section in the main paper.

\begin{table*}
    \centering
    \begin{tabular}{c|ccccccccc}
    \toprule
       Metrics  & Chair & Drums & Ficus & Hotdog & Lego & Materials & Mic & Ship & Overall\\
    \hline
       PSNR  & 35.7522 & 34.3105 & 32.3541 & 34.8191 & 32.9918  & 34.4048 & 35.1800 & 32.2207 &34.0042\\
       SSIM  & 0.9413 & 0.9283 & 0.9161 & 0.9219 & 0.9081  & 0.9440 & 0.9421 & 0.9535 &0.9319\\
       LPIPS  & 0.0083 & 0.0019 & 0.0016 & 0.0029 & 0.0013  & 0.0076 & 0.0037 & 0.0009 &0.0035\\
       Bit Acc  & 95.22\% & 94.98\% & 94.21\% & 91.69\%  & 92.56\% & 95.65\% & 90.52\% & 92.89\% &93.46\%\\
    \bottomrule
    \end{tabular}
    \caption{Additional results of GaussianSeal on NeRF Synthetic dataset.}
    \label{tab:sync}
\end{table*}

\subsection{PyTorch-like Pseudo Code of Adaptive Bit Modulation}
We here provide PyTorch-like pseudo code for adaptive bit modulation modules definition, and how they are applied in LGM inference process.
\begin{algorithm*}[htp]
\caption{Adaptive Bit Modulation}

\begin{minted}{python}
import torch.nn as nn
class View(nn.Module):
    def forward(self, x):
        return x.view(*self.shape)
# define modulation modules
# input
noise_block_size = 8
message_len = 16
b_in = nn.Sequential(
    nn.Linear(message_len, 4 * noise_block_size * noise_block_size), nn.SiLU(),
    View(-1, 4 , noise_block_size, noise_block_size), )
b_in_conv = torch.nn.Conv2d(4, 4, 3, padding=1)
# middle
b_mid = nn.Sequential(
    nn.Linear(message_len, 512 * noise_block_size * noise_block_size), nn.SiLU(),
    View(-1, 512 , noise_block_size, noise_block_size), )
b_mid_conv = torch.nn.Conv2d(512,512,3,padding=1)
# output
b_out = nn.Sequential(
    nn.Linear(message_len, 512 * noise_block_size * noise_block_size), nn.SiLU(),
    View(-1, 512 , noise_block_size, noise_block_size), )
b_out_conv = torch.nn.Conv2d(512,512,3,padding=1)
# define adaptive coefficients
watermark_alpha = nn.Parameter(torch.tensor(0.1))
watermark_beta = nn.Parameter(torch.tensor(0.1))
watermark_gamma = nn.Parameter(torch.tensor(0.1))
# forward process for UNet with modulation
def forward_with_watermark(x, m): # x is input image, m is message
    H, W, B, C = x.shape[2], x.shape[3], x.shape[0], x.shape[1]
    # modulation in input stage
    m_in = b_in(m).repeat(B,1,int(H/noise_block_size),int(W/noise_block_size))
    m_in = b_in_conv(m).repeat(1,3,1,1)[:, :9, :, :]
    x = conv_in(x)
    x = x + watermark_alpha * m_in
    # encoder
    x = unet.encoder(x)
    H,W = x.shape[2], x.shape[3]
    # modulation in middle
    m_mid = b_mid(m).repeat(B,1,int(H/noise_block_size),int(W/noise_block_size))
    repeat_times = int(x.shape[1]/m_mid.shape[1])
    m_mid = b_mid_conv(m_mid).repeat(1,repeat_times,1,1)
    x = x + watermark_beta * m_mid
    # decoder
    x = unet.decoder(x)
    H, W = x.shape[2], x.shape[3]
    m_mid = b_out(m).repeat(B,1,int(H/noise_block_size),int(W/noise_block_size))
    m_mid = b_out_conv(m_mid)[:, :x.shape[1], :, :]
    x = x + watermark_gamma * m_mid
    
    # conv out to get output 3DGS tensor
    x = conv_out(x)
    return x
\end{minted}

\end{algorithm*}

\section{Additional Experiments}
\label{sec: expr}

\subsection{Evaluation on NeRF Synthetic Dataset}
\begin{figure*}
    \centering
    \includegraphics[width=0.5\linewidth]{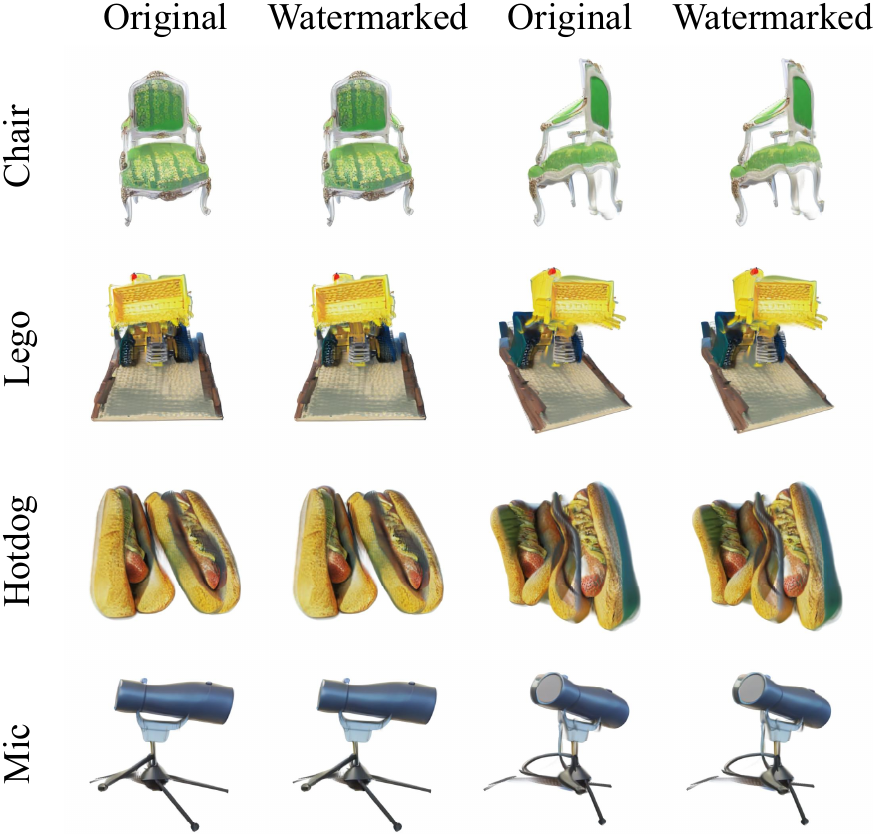}
    \caption{Visualized results of objects from NeRF Synthetic dataset, generated via original LGM model and watermarked LGM model.}
    \label{fig:nerf}
\end{figure*}

To evaluate the generalization ability of our proposed GaussianSeal, we leverage the bit modulation trained on the Objaverse dataset and test on the NeRF Synthetic dataset which is not used during the training process. Details of dataset settings can be referred from Sec.~\ref{sec: dataset_details}. Quantitative results are shown in Tab.~\ref{tab:sync}, and qualitative results are shown in Fig.~\ref{fig:nerf}.

\subsection{Evaluation of Generalization and Scalability}
In the main experiments, we randomly select 10,000 and 100 objects separately from Objaverse dataset for training and validation. These two subsets \textbf{do not have overlap}. To validate the generalization and scalability of GaussisanSeal, we have done the validation on 10,000 and 100,000 objects from Objaverse dataset, which also have no overlap with training and validation datasets. The results are as below in Tab.~\ref{tab:scale}, which shows that our method has good \textbf{scalability and generalizability} on larger unseen datasets.
% \vspace{-2.7mm}
\begin{table}[h]
    \centering
    % \vspace{-3mm}
    % \resizebox{\linewidth}{!}{
    \begin{tabular}{c|cccc}
        \hline
        Amount & PSNR & SSIM & LPIPS & Bit Acc \\ \hline
        10,000   & 37.4815   & 0.9705   & 0.0038   & 96.41\%   \\ \hline
        100,000   & 37.0329   & 0.9639   & 0.0056   & 96.27\%  \\ \hline
    \end{tabular}
    % }
    \label{tab:generalize}
    % \vspace{-2.5mm}
    \caption{Results of larger-scale validation.}
    \label{tab:scale}
    % \vspace{-3mm}
\end{table}

\subsection{Watermark Capacity of GaussianSeal}
In the experiments section of the main paper, we hide 16 and 32 bits in the 3D generation model, and here, we try to hide more bits to explore the upper limit of the bit watermarking capacity of our method. We extend message length to 48 and 64, and provide the quantitative results in Tab.~\ref{tab:capacity}. 
% \begin{table}[h!]
% \centering
% \begin{tabular}{|c|c|c|}
% \hline
% \multirow{2}{*}{c1} & 1 & \multirow{2}{*}{c3} \\ \cline{2-2} 
%                         & 2 &                        \\ \hline
% \end{tabular}
% \caption{1}
% \end{table}
\begin{table*}
    \centering
    \begin{tabular}{c|c|cccc}
    \toprule
       Message Length & Method & PSNR & SSIM & LPIPS & Bit Acc \\
    \hline
       \multirow{2}{*}{48} & GaussianMarker & 32.8155 & 0.9373 & 0.0088 & 91.14\%\\
                           & GaussianSeal (Ours) & 33.0690 & 0.9463 & 0.0078 & 91.67\%\\
    \hline
       \multirow{2}{*}{64} & GaussianMarker & 30.9979 & 0.9053 & 0.0061 & 90.34\%\\
                           & GaussianSeal (Ours) & 31.3665 & 0.9267 & 0.0045 & 92.77\%\\
    \bottomrule
    \end{tabular}
    \caption{Watermark capacity exploration of our method, compared to current 3DGS watermarking state-of-the-art method GaussianMarker.}
    \label{tab:capacity}
\end{table*}

\subsection{Results on novel views}
We supply experiments on novel views here. Results are shown in Tab.~\ref{tab:novel}, which show that our method is \textbf{generalizable} to novel views.
% \vspace{-0.9mm}
\begin{table}
    \centering
    % \vspace{-5mm}
    % \resizebox{\linewidth}{!}{
    \begin{tabular}{ccccc}
    \hline
        PSNR & SSIM & LPIPS & Bit Acc\\
    \hline
        32.2195  & 0.9617 & 0.0060 & 96.61\%\\
    \hline
    \end{tabular}
    % }
    % \vspace{-3mm}
    \caption{Results of novel views unseen in training.}
    \label{tab:novel}
    % \vspace{-3mm}
\end{table}

\subsection{Robustness against 3DGS compressing}
We adopt C3DGS~\cite{lee2024compact} to the generated 3DGS objects, which are compressed by 25 times. Results shown below in Tab.~\ref{tab:compress}.
% \vspace{-0.5mm}
\begin{table}[h]
    \centering
    % \vspace{-4.5mm}
    % \resizebox{\linewidth}{!}{
    \begin{tabular}{cccc}
    \hline
         PSNR  & SSIM & LPIPS & Bit Acc\\
    \hline
         35.6889  & 0.9458 & 0.0089 & 93.51\%\\
    \hline
    \end{tabular}
    % }
    % \vspace{-5mm}
    \caption{Results of robustness validation on 3DGS compression.}
    \label{tab:compress}
    % \vspace{-7mm}
\end{table}

\subsection{Robutness against fine-tuning attack}
% 在我们的threat model中，我们设想的使用场景为模型拥有者提供类似API的服务，用户和攻击者只能输入图像，并得到对应的添加水印的3DGS模型，无法获取模型的结构和权重。但考虑到安全需求，我们在表x中展示了模型对轻度微调攻击的鲁棒性。微调攻击的形式是，采用不含bit loss的损失函数，以此微调LGM部分网络的权重，降低GaussianSeal的水印性能。
In our threat model, we envision a scenario where the model owner provides an API-like service, allowing users and attackers to input images and receive a watermarked 3DGS model, but without access to the model's structure or weights. To address security concerns, we demonstrate the model's robustness against mild fine-tuning attacks in Tab.~\ref{tab:finetune}. Such attack involves fine-tuning the LGM network weights using a loss function that excludes bit loss, which reduces the watermark performance of GaussianSeal.

\begin{table}
    \centering
    \begin{tabular}{c|c}
    \toprule
        Steps &  Bit Acc\\
    \hline
        20 &  96.88\%\\
        40 &  94.53\%\\
        60 &  92.97\%\\
        80 &  91.26\%\\
        100 &  89.84\%\\
    \bottomrule
    \end{tabular}
    \caption{Results of robutsness against fine-tuning attack.}
    \label{tab:finetune}
\end{table}

\subsection{Results of Security Analysis}
\begin{figure*}[t!]
	\centering
	\includegraphics[width=0.5\linewidth]{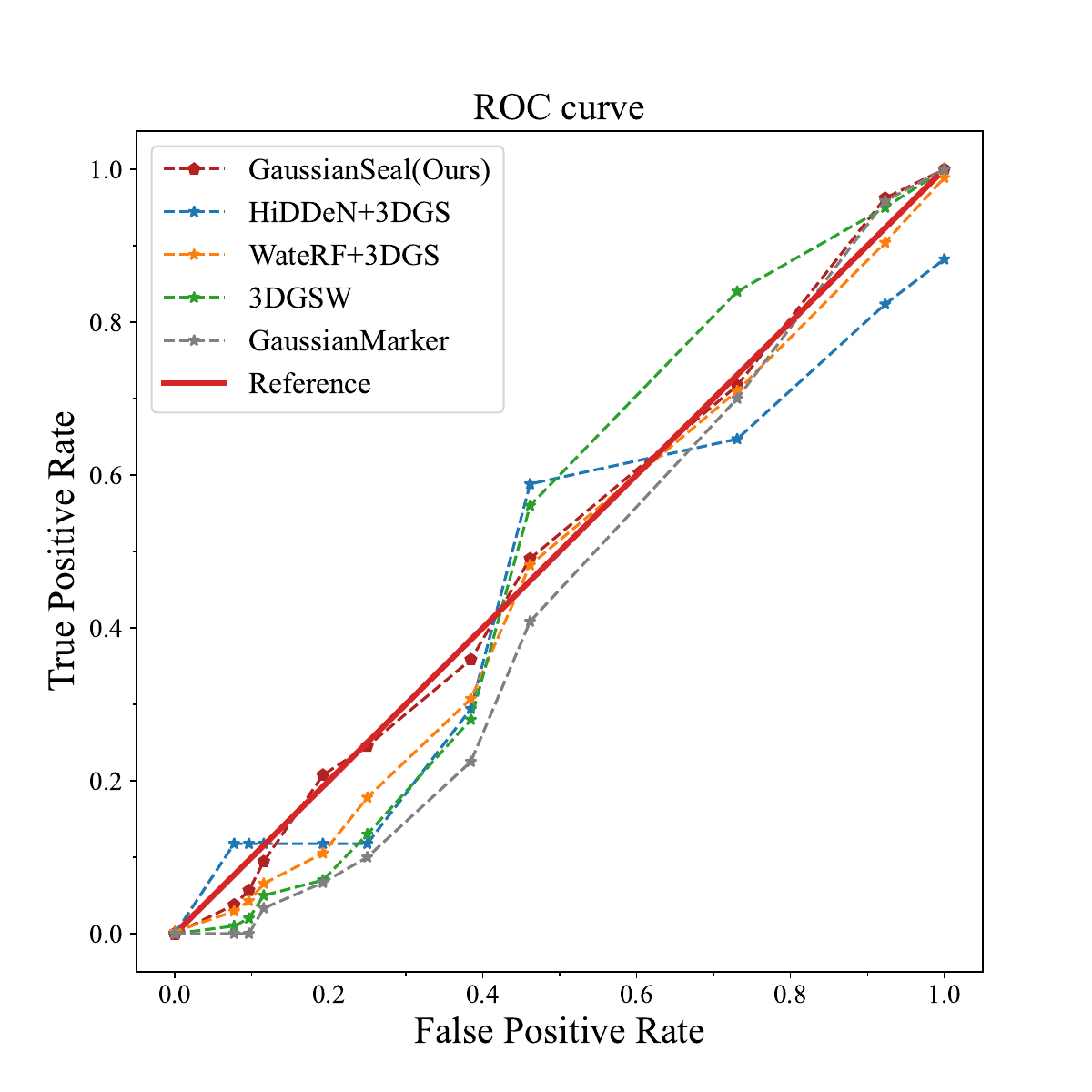}
    \vspace{-15pt}
	\caption{The ROC curve of various methods under a steganography detector. A curve closer to the reference central axis (indicating random guessing) signifies better security for the corresponding method.}
 \vspace{-5pt}
    \label{roc}
\end{figure*}

To evaluate the security of our GaussianSeal, we conduct anti-steganography detection using StegExpose~\cite{boehm2014stegexpose} on container images generated by various methods, including 3D-GSW~\cite{jang20243d}, GaussianMarker~\cite{huang2024gaussianmarker}, 3DGS+HiDDeN~\cite{zhu2018hidden}, 3DGS+WateRF~\cite{jang2024waterf}, and our GaussianSeal.  
Each method embeds 16 bits into 3DGS objects. We adjust the detection thresholds across a wide range, from 0.001 to 0.95, within StegExpose~\cite{boehm2014stegexpose}, and plot the resulting receiver operating characteristic (ROC) curve in Fig.~\ref{roc}. The ideal scenario assumes the detector has a 50\% chance of identifying a watermark in a balanced test set, equivalent to random guessing.  
The results clearly demonstrate that our method significantly outperforms 3DGS watermarking methods in terms of security. This indicates that our approach is much less susceptible to detection by steganography analysis techniques, thereby achieving a robust level of security.

\section{Discussion and Analysis}
\label{sec: discuss}

\subsection{GaussianSeal with 3D feature}
Our GaussianSeal method draws on some watermarking practices used in Text-to-Image (T2I) models and incorporates specialized designs for the representation of 3D objects and scenes, as well as properties unique to 3DGS, to avoid compromising the generation quality of 3D models. Specifically, our considerations and designs are as follows:

\noindent \ding{113}~We observe that the attributes of 3DGS are highly sensitive to changes in values; even minor parameter adjustments can cause significant degradation in the visual quality of 3DGS point clouds, resulting in issues such as Gaussian point dispersion and irregular deformation. Based on these observations, we (1) embed the watermark into the output of the UNet rather than into the final 3DGS tensor used for representing 3D objects, and (2) introduce an adaptive coefficient for the embedded watermark tensor to minimize its impact on the original model output values. Similarly, we opt for bit modulation of the watermark instead of directly fine-tuning the UNet, as shown in the main paper, where direct fine-tuning of the UNet leads to artifacts in the generated 3DGS results.

\noindent \ding{113}~For 3DGS generative models, such as LGM, although they are similar to commonly used diffusion models in image generation, directly applying diffusion watermarking methods such as fine-tuning, designing LoRA, or adapter-based methods still results in reduced model quality. This is because the latent code in diffusion models is trained through a noise-adding and denoising process, making the UNet robust to noise in the latent code and capable of producing good outputs even when disturbances occur. However, in 3D generative models like LGM, the training does not involve robustness to noise, so changes in UNet weights and intermediate results between blocks lead to significant changes in the generated output. Therefore, a better approach for embedding bit information is needed. Through experiments, we find that using an additional lightweight modulation module to embed bit information at key locations achieves a better trade-off between precise bit decoding and maintaining generation quality.

\noindent \ding{113}~Regarding watermark extraction, previous studies on watermarking methods for 3D objects and scenes often choose to decode from the DWT low-low subband of rendered results to improve decoding accuracy. We adopt this mechanism in our watermarking of 3D generative models, and experiments show that such an approach is also effective in 3D generation model watermarking.
\subsection{Relationship to other generation model watermarking methods}
First, we define the terms \textit{post-generation} and \textit{in-generation} watermark methods mentioned in the main paper. The \textit{post-generation} watermarking method refers to the technique of adding watermarks to the generated results of a 3D Gaussian generation \textbf{after} the model produce its output. Since this method is a post-processing step and is independent and decoupled from the specific generation process of the model, it is termed post-generation. This approach requires adding watermarks to each generated result individually, thus necessitating a longer time and greater consumption of computational resources.

The \textit{in-generation} watermarking method refers to the process where watermarks are embedded \textbf{during} the generation process of the model. This method is closely coupled and tightly integrated with the generation model, such that the results produced by the model already contain the watermarks, eliminating the need for any post-processing. The advantage of this method is that with a single training session, one can directly obtain the target with the watermarks embedded.

We compare our GaussianSeal with existing in-generation T2I generation model watermarking methods which are similar to ours, highlighting the differences in approach and implementation between the proposed GaussianSeal and the methods mentioned below.

\noindent \ding{113}~\textbf{Compared to AquaLoRA}: AquaLoRA~\cite{feng2024aqualora} fine-tunes the UNet of Stable Diffusion (SD) using LoRA to enable the extraction of bit information from the generated results. This method requires substantial computational resources and extensive training periods—our previous attempts indicated that it necessitates over 72 hours of training on an A100 GPU—and it can adversely affect the quality of the model generation. This approach is markedly different from our proposed solution.

\noindent \ding{113}~\textbf{Compared to LaWa}: LaWa~\cite{rezaei2024lawa} achieves precise bit decoding by adding a bit embedding module to the decoder of SD and training it. In contrast, our method opts to incorporate a bit modulation mechanism within the UNet, and considering the sensitivity of 3DGS to weight variations, we have introduced adaptive value range compression to prevent the generation of poor-quality results.

\noindent \ding{113}~\textbf{Compared to WaDiff}: WaDiff~\cite{min2024watermark} decodes bits by fine-tuning the first layer of SD UNet and training an additional bit embedding layer. Our approach, on the other hand, completely freezes the original generation model, which is a consideration aimed at addressing the fragility of LGM model weights.

In short, our method exhibits distinct differences in implementation and conceptual approach compared to previous watermarking techniques for generative models. These differences stem from our observations and considerations of the properties of 3DGS and its generative models.

\section{More Visualized Results}
\begin{figure*}
    \centering
    \includegraphics[width=0.7\linewidth]{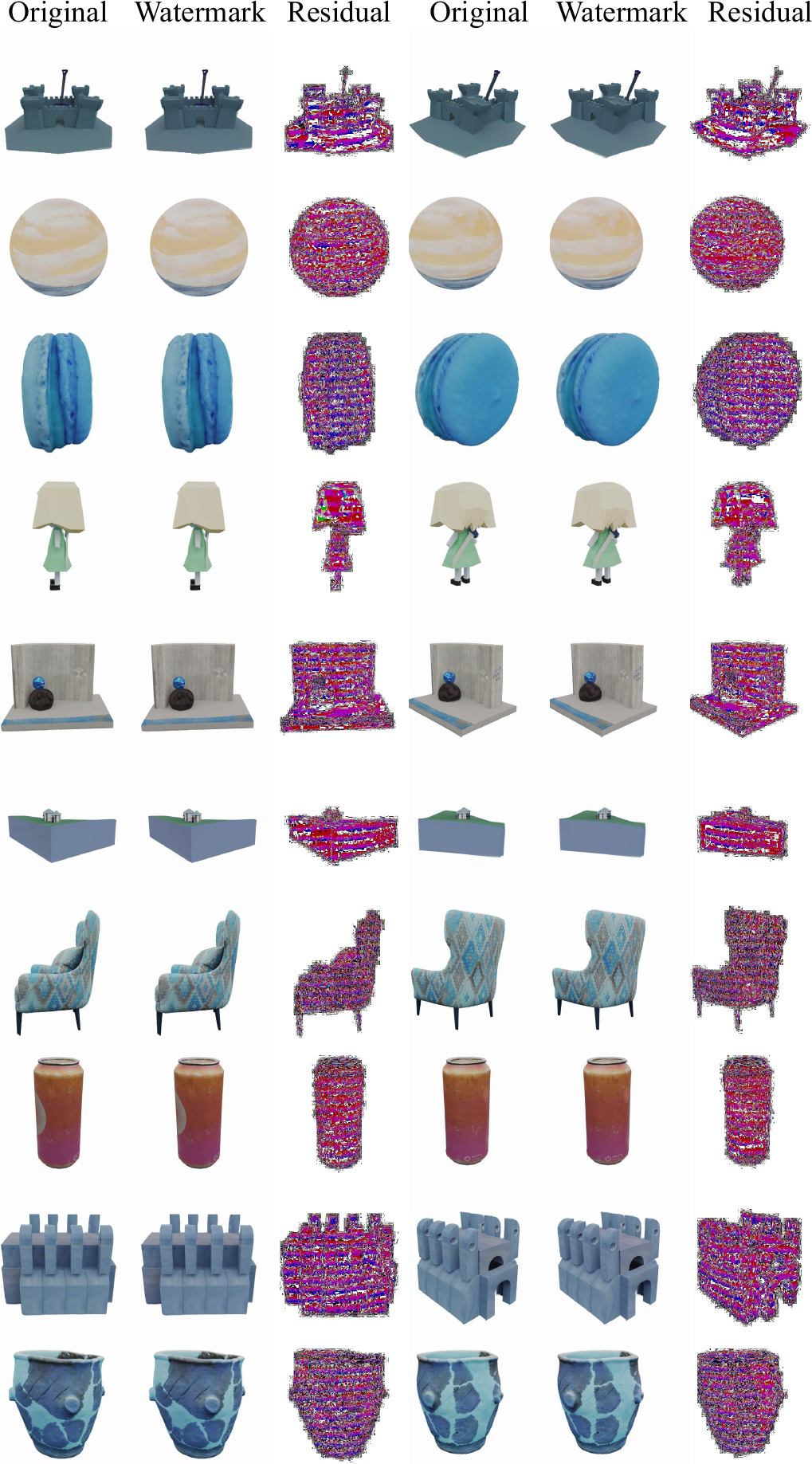}
    \caption{More visualized results of GaussianSeal, each object with two views.}
    \label{fig:more}
\end{figure*}

Here we provide more visualized results of 3DGS objects watermarked by GaussianSeal in Fig.~\ref{fig:more}, including original generated object, watermarked object, and their residual image.
\label{sec: visual}
\clearpage
{
\small
\bibliographystyle{ieeenat_fullname}
\bibliography{main}
}


\end{document}